\documentclass[10pt,journal,compsoc]{IEEEtran}

\ifCLASSOPTIONcompsoc
  \usepackage[nocompress]{cite}
\else
  \usepackage{cite}
\fi
\ifCLASSINFOpdf
\else
\fi
\usepackage{xcolor,soul,framed} %,caption

\colorlet{shadecolor}{yellow}
\usepackage[pdftex]{graphicx}
\graphicspath{{../pdf/}{../jpeg/}}
\DeclareGraphicsExtensions{.pdf,.jpeg,.png}

\usepackage[cmex10]{amsmath}
\usepackage{array}
\usepackage{mdwmath}
\usepackage{mdwtab}
\usepackage{eqparbox}
\usepackage{url}
\usepackage{hyperref}
\usepackage{tabularx}
\usepackage{subcaption}
\hypersetup{citecolor=Green}
\usepackage{mathrsfs}
\usepackage{rsfso}
\usepackage{amssymb}
\usepackage{multirow}
\usepackage{graphicx}
\usepackage{fancyhdr}

\begin{document}
\title{Learning Deformable 3D Graph Similarity to Track Plant Cells in Unregistered Time Lapse Images}

\author{Md Shazid Islam,~\IEEEmembership{}
        Arindam Dutta,~\IEEEmembership{}
        Calvin-Khang Ta,~\IEEEmembership{}
        Kevin Rodriguez,~\IEEEmembership{}
        Christian Michael, ~\IEEEmembership{}
        Mark Alber,~\IEEEmembership{}
        G. Venugopala Reddy,~\IEEEmembership{}
        Amit K. Roy-Chowdhury,~\IEEEmembership{Fellow,~IEEE}% <-this % stops a space

\IEEEcompsocitemizethanks{\IEEEcompsocthanksitem Md Shazid Islam is with the Department
of Electrical and Computer Engineering, University of California Riverside, Riverside, California, USA-92507.\protect\\
E-mail: misla048@ucr.edu
% \IEEEcompsocthanksitem.
}% <-this % stops a space
% \thanks{Manuscript received July 19, 2023; revised August 26, 2023.}
}
% a space would be appended to the last name and could cause every nahttps://www.overleaf.com/project/638aa78e90bd0b2d9ad18e75me on that


%%%%%%%%%

\markboth{IEEE/ACM Transactions on Computational Biology and Bioinformatics}%
{Shell \MakeLowercase{\textit{et al.}}: Plant Cell Tracking}

%%%%%%%%%





\IEEEtitleabstractindextext{%
\begin{abstract}
Tracking of plant cells in images obtained by microscope is a challenging problem due to biological phenomena such as  large number of cells, non-uniform growth of different layers of the tightly packed plant cells and cell division. Moreover, images in deeper layers of the tissue being noisy and unavoidable systemic errors inherent in the imaging process further complicates the problem. In this paper, we propose a novel learning-based method that exploits the tightly packed three-dimensional cell structure of plant cells to create a three-dimensional graph in order to perform accurate cell tracking. We further propose novel algorithms for cell division detection and effective three-dimensional registration, which improve upon the state-of-the-art algorithms. We demonstrate the efficacy of our algorithm in terms of tracking accuracy and inference-time on a benchmark dataset.
\end{abstract}

}

\maketitle

% is not selected <OR> if conference mode is selected - because compsoc
% conference papers position the abstract like regular (non-compsoc)
\IEEEdisplaynontitleabstractindextext

% \fi
\IEEEpeerreviewmaketitle

\ifCLASSOPTIONcompsoc
\IEEEraisesectionheading{\section{Introduction}\label{sec:introduction}}
\else
\section{Introduction}
\label{sec:introduction}
\fi

\IEEEPARstart{M}{\lowercase{orphogenesis}} analysis is a cardinal topic of interest in computational biology, which analyzes the development of various biological forms including cell growth and division patterns for both plant and animal tissues. Tracking the development of these cells over time provides an excellent description of several physiological properties such as structural integrity \cite{orynbayeva2015metabolic}, enzyme activity \cite{fry2004primary}, gene expression \cite{shen2006automated}, cell division \cite{liu2009endocytic} amongst several others. 

\begin{figure*}
  \centering
  \includegraphics[width =  \textwidth]{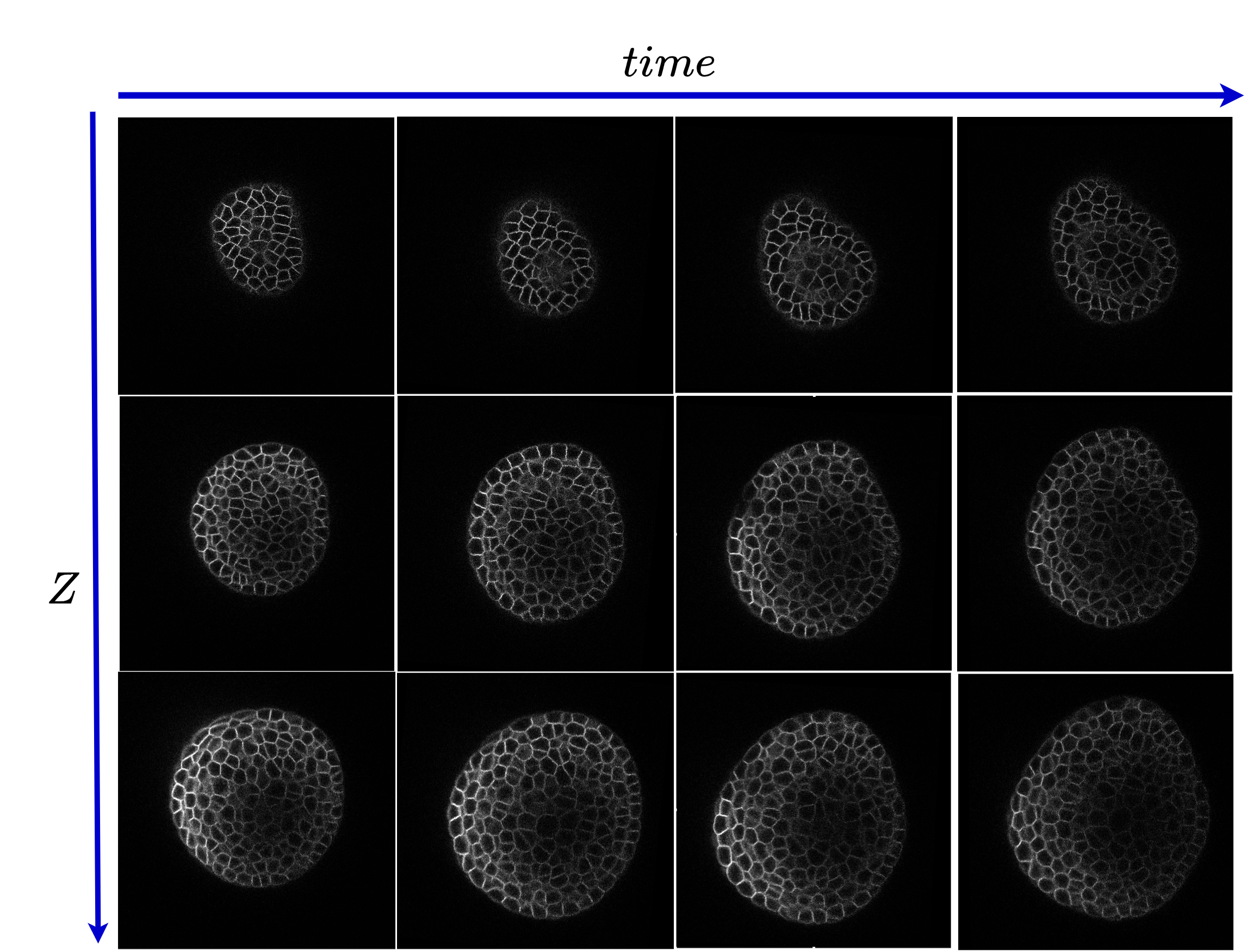}
  \caption{The time series of microscopic plant cell images are shown. There is a time axis and a Z-axis. Z-axis which goes along the depth of the plant indicates the slice number and the time axis indicates the time passed during imaging. According to the dataset \cite{willis2016cell}, time difference between two time points is 4 hours. On Z-axis, images of 3 slices are shown. Those slices are sampled from the top, middle, and bottom of the SAM, respectively.}
  \label{cell_t_z}
\end{figure*}

The plant of interest for this study is  {\it Arabidopsis Thaliana} \cite{refahi2021multiscale}. This plant is purposefully chosen as its morphological structure resembles several other plants. Thus, its study aids in understanding the morphological properties of several other plants as discussed by the authors of \cite{koch2008diversity, zhang2007signalling}. In order to analyze the cell growth and division characteristics, this work focuses on the Shoot Apical Meristems (SAM) \cite{shani2006role} of the plant. Shoot Apical Meristems (SAM) are densely packed multi-layer tissues that provides cells for leaves, stems, and branches. With the advancement of microscopy imaging techniques, time-lapse images of long time intervals can be collected from SAM by virtue of Confocal Laser Scanning Microscopy (CLSM) \cite{paddock1999confocal} based live cell imaging. In this imaging technique, a laser beam is moved along the depth of the plant which results in a series of two-dimensional images being captured. The two-dimensional images are referred to as {\it slices} which, when put together forms the image {\it stack} of the plant. Fig. \ref{cell_t_z} shows some exemplar slices of the SAM region of {\it Arabidopsis Thaliana} captured by CLSM.

\begin{figure}
\centering
    \subfloat[] {{\includegraphics[width= 0.6 \columnwidth]{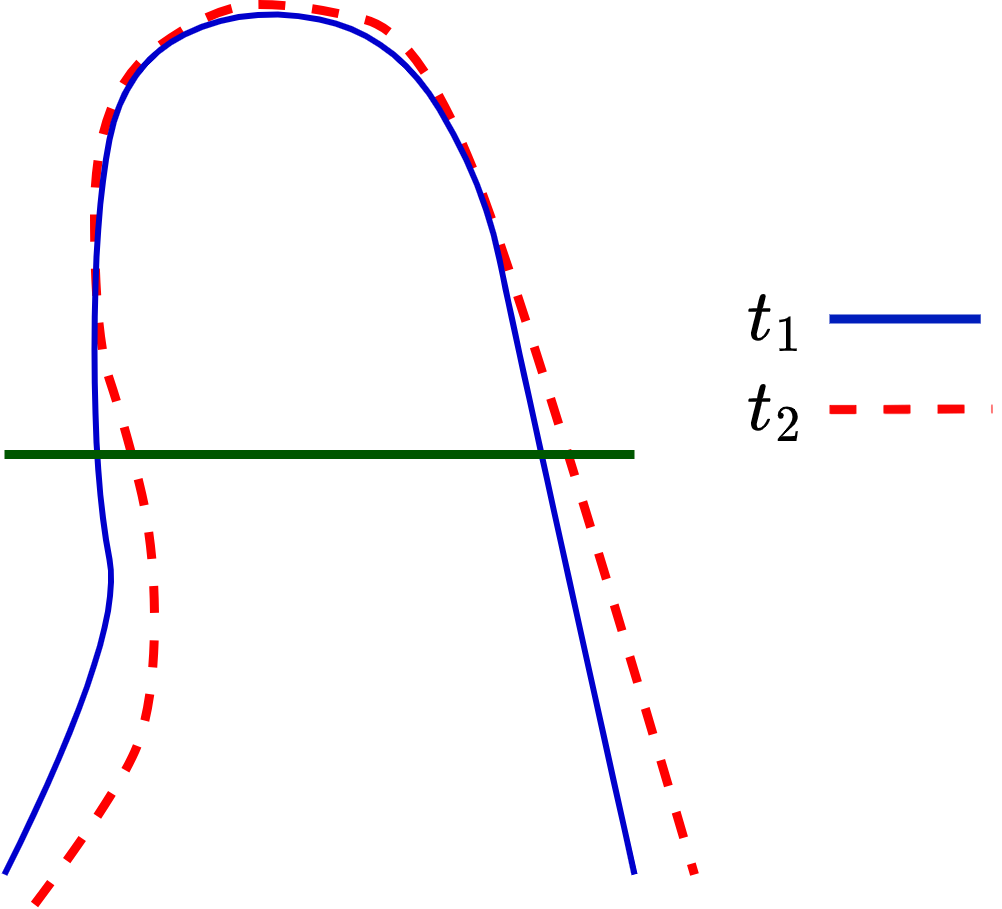}} \label{tilt_1}}
    
    \subfloat[]{{\includegraphics[width= \columnwidth]{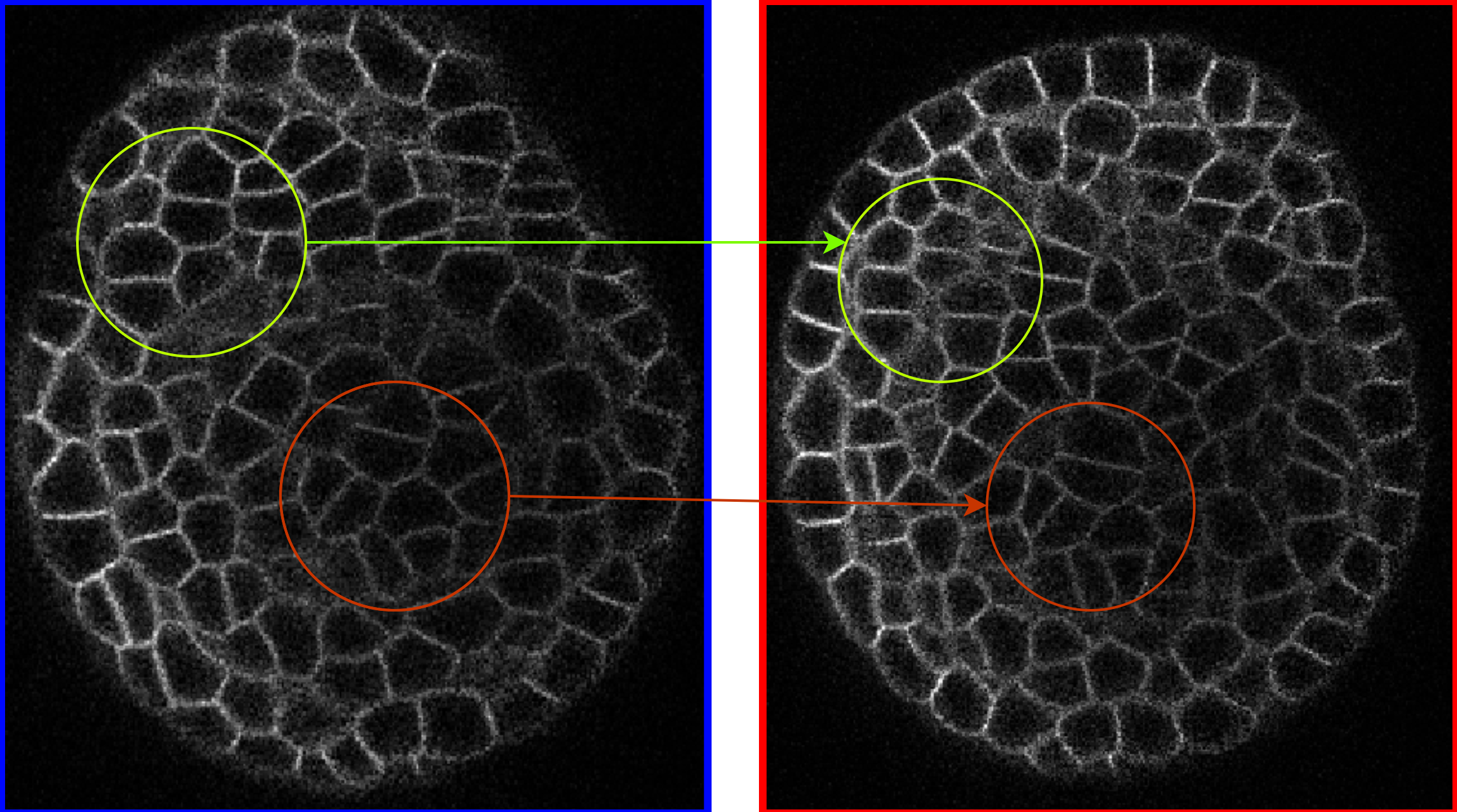}}\label{tilt_2}} 
    
    \caption{(a) the overlay of SAM in two time-points when tilting occurs. The blue solid and red dotted lines indicate time $t_{1}$ and $t_{2}$, respectively. (b) The cross-section view with respect to the green line in (a) is shown. The marked patches reveal that the shape and size of the same cells captured in the two-dimensional plane change due to tilt.}\label{tilt}
\end{figure}

\begin{figure*}[!htb]
  \centering
  \includegraphics[width = \textwidth]{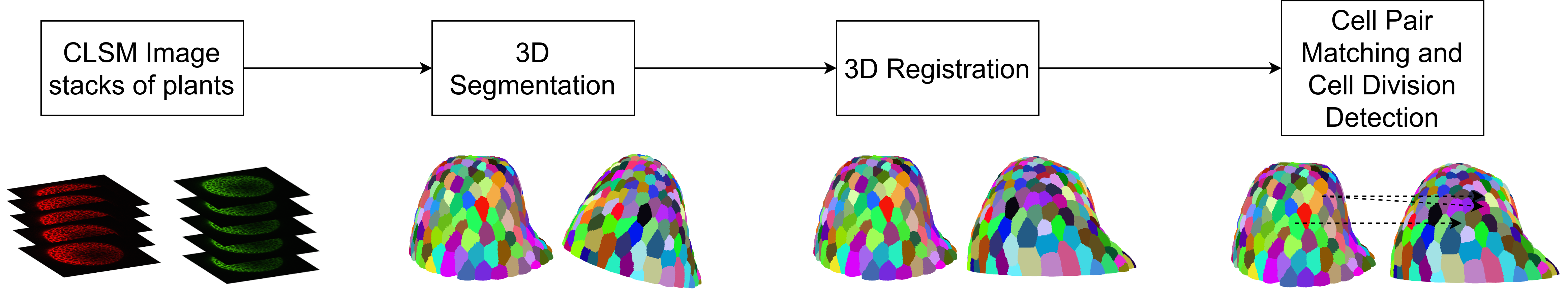}
  \caption{The entire workflow is shown in the figure. Image stacks obtained by CLSM imaging are segmented using a 3D segmentation technique. Then registration is done for pairwise time points using our proposed 3D registration method. Finally, our novel learning-based approach is used for cell pair matching and cell division detection.}
   \label{workflow}
\end{figure*}
The Shoot Apical Meristems (SAM) of {\it Arabidopsis Thaliana} is a tightly-packed structure with a large number of cells. Spatio-temporal tracking of such large number of cells is both computationally expensive and time consuming. In addition, the images captured from the deep layers of SAM suffer from low signal to noise ratio (SNR), as mentioned in \cite{9153414}, which makes the problem of tracking of deep-layer cells very challenging. Further, plant has a natural tendency of growing towards the light source it gets exposed to \cite{michael2013intelligent}. As a result, when a plant is kept in a non-uniform lighting setup, the plant leans toward the light source and gets tilted. The live cell imaging technique can not capture this information as it only captures two-dimensional images on X-Y plane (or, top view), thus further complicating the problem of tracking of plant cells. Hence, there's a need for an automated, fast, and robust algorithm that can perform tracking of plant cells in Shoot Apical Meristems (SAM) of {\it Arabidopsis Thaliana}.

There are several works on plant cell tracking. Most of them are based on two dimensional local-graph matching \cite{jiang2019cnn, liu2019deepseed, liu2009robust, liu2016robust} which exploits the tight spatial topology of neighboring cells. In these works, two-dimensional local star graph \cite{mendia1992optimal} is constructed by connecting the centroid of the cell of consideration to the centroids of its neighboring cells. This graph structure represents the neighboring structure of the cell of consideration. Two cells from different time points which have the closest graph structure are considered as the corresponding cells (also known as matching cells). In these works,  cell correspondences  are done between a pair of slices of two consecutive time points, which are then combined to obtain pairwise tracks over the entire stack.  Chakraborty {\it et al.}  \cite{chakraborty2015context} proposed a conditional random field (CRF) \cite{sutton2012introduction} based approach where all cells present on a slice are considered as nodes of the graph and all cells having common boundary are connected to form the graph. Marginal posteriors of each node are calculated using loopy belief propagation  \cite{ihler2005loopy} followed by a graph labeling method to obtain the optimal correspondence. However, this method is not scalable for larger datasets as the graph labeling optimization step is fairly slow. A few recent works \cite{liu2017cell,liu2018multi} form three-dimensional local graph of a cell by combining  two-dimensional local star graphs of adjacent slices. However, these approaches do not use the inter-slice
the connection among the cells
, thus ignore the three-dimensional spatial information of the  cells. In addition, these works make an assumption that two matching cells must have the same number of neighbors which makes these approaches very susceptible to segmentation error. Another recent method, DeepSeed \cite{liu2019deepseed} uses the weighted sum of shape similarity and neighboring structure similarity scores between the cells. The correspondent cell pairs (also termed as seed pairs) are then selected on the basis of a similarity score threshold. This is followed by using relative position with respect to the seed pairs, to track rest of the cells in an iterative fashion. Instead of using the shape similarity of individual cell,  \cite{deep_patch} compares the patches of cells of two different time points and uses the K-M algorithm \cite{KM}, for patch association. However, these methods are highly dependent on segmentation accuracy and hence result in subpar performance for deeper layer cells of the plant. 

Another limitation of the algorithms \cite{chakraborty2015context, liu2019deepseed, deep_patch} present in the literature is solving tracking problem only in a two-dimensional plane. This assumption ignores phenomena like the tilting of plants. Fig \ref{tilt} illustrates that when tilt occurs  the same cells exhibits a change of shape or size when cross-section is observed. As a result, the methods which are largely dependent on two-dimensional cell shape in tracking fail under tilting conditions of the plant. We address this issue by considering the entire three-dimensional structure of the plant.

In order to address the aforementioned problems, we make the following contributions in this paper:

 \begin{itemize}
     \item We develop a novel learning-based tracking algorithm for plant cells that uses three-dimensional geometric information of the tightly packed plant cell structure. 
     
     \item We propose a three-dimensional graph matching technique where graphs are connected to $k$ nearest neighbors to extract contextual information of the cell structure resulting in increased tracking accuracy. 

     \item We further develop a novel learning-based cell division detection technique which uses both the three-dimensional shape and local graph similarity to detect mother and daughter pairs.

     \item Finally, we show that our method is inference friendly, with $\approx 12 \times$ improvements in terms of inference time against the method presented in \cite{chakraborty2015context}.\\

\end{itemize} 
The organization of this paper is as follows: 
section \ref{sec:methodology} provides a detailed description of the  Methodology. Experiments and results are discussed in section \ref{sec:Experiment}.  Finally, section \ref{sec:Conclusion} concludes the paper.

\begin{figure*}[!htb]
\centering
    \subfloat[]{{\includegraphics[width=0.2\textwidth]{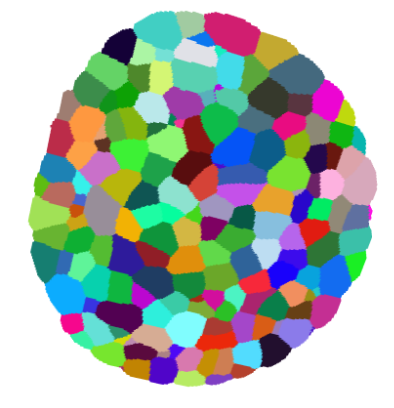}} \label{horizontal_section}}
    \hfill
    \subfloat[]{{\includegraphics[width=0.3\textwidth]{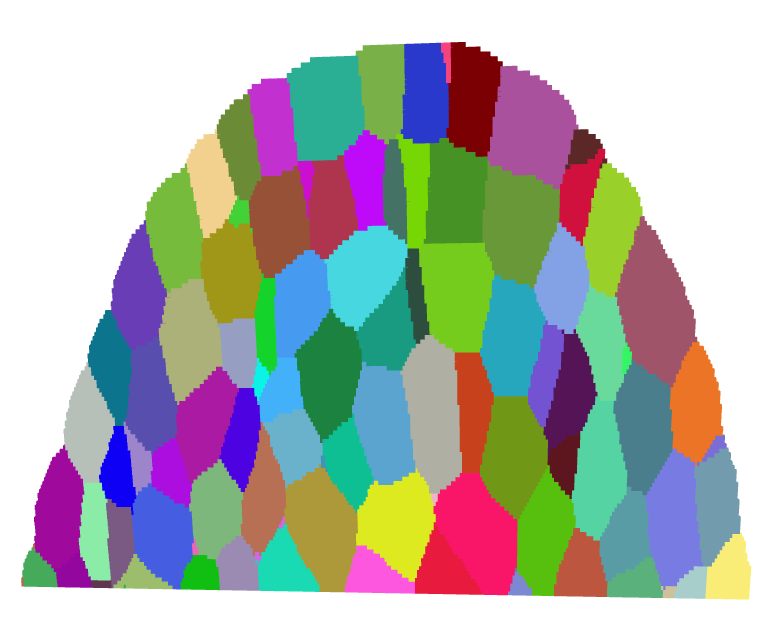}}\label{vertical_section}}
    \hfill
    \subfloat[]{{\includegraphics[width=0.3\textwidth]{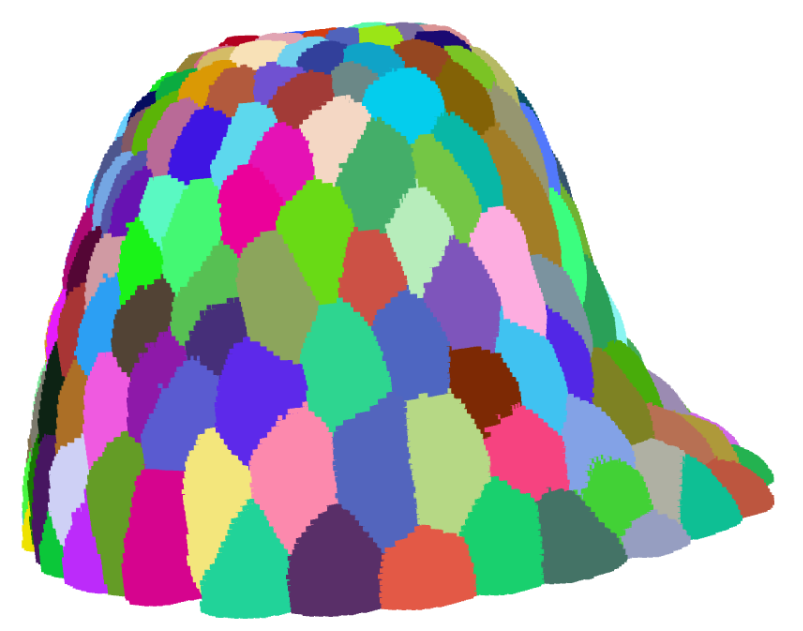}}\label{3D_recon}}
    % \hfill
    \caption{Segmentation of plant cells using Cellpose3D. (a) Cell segmentation in a horizontal slice. (b) Cell segmentation on the vertical slice.  (c) 3D reconstruction using the 3D instance segmentation}\label{segmentation}
\end{figure*}

\begin{figure*}[!htb]
  \centering
  \includegraphics[width =  \textwidth]{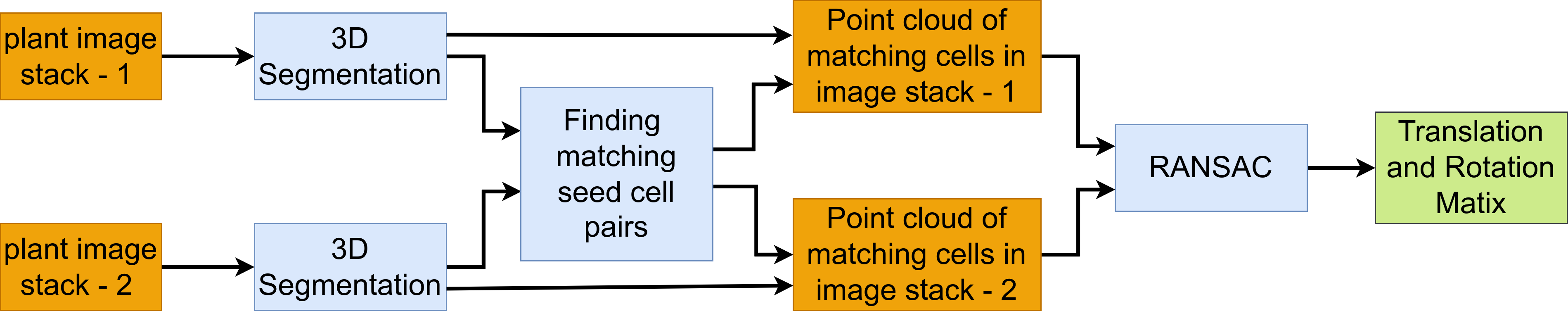}
  \caption{Proposed  method for registration of plant stacks. At first the three-dimensional instance segmentation are obtained using Cellpose3D. Then a number of correspondent cell pairs (seed pairs) are determined using the method described in \cite{liu2019deepseed}. The point cloud of seed pair cells of each stack are extracted. Using the FPFH feature of the point cloud, RANSAC algorithm is used to determine the translation and rotation matrix which is used for registration.}
  \label{registration}
\end{figure*}

\section{Methodology}
\label{sec:methodology}

The core problem that this paper tackles is tracking plant cells over a long time period. The inputs are unregistered plant image stacks of multiple time points with a constant time interval. Over this time period, some cells get divided. The overall objective is to  track tightly packed plant cells and detect cell division events in order to determine long-period cell lineages. In this section, we elaborate on our proposed pipeline for performing robust tracking of plant cells.  Firstly, three-dimensional segmentation of CLSM plant stack images is done. Then, three-dimensional registration is involved to align the plant image stacks over time. Finally, a novel learning-based three-dimensional local graph matching technique is employed to find cell pair matches of tightly packed plant cells and detect cell division events. Fig. \ref{workflow} shows the entire workflow of our proposed algorithm.

\subsection{Segmentation}
 Cell segmentation is the first step for cell-tracking algorithms which helps identify individual instances of all cells, present in the plant. State-of-the-art cell tracking methods such as \cite{liu2019deepseed,deep_patch,chakraborty2015context}  focused on watershed segmentation \cite{beucher1993morphological} technique that segments the cell border on every slice. However, watershed segmentation does not provide the three-dimensional spatial relationships of the cells among the slices. Recently, learning-based techniques such as spherical harmonics \cite{eschweiler2021spherical}, U-Net watershed \cite{eschweiler2019cnn} , Cellpose3D \cite{eschweiler2022robust} have been proposed, which provide instances of every cell in three-dimensional space. Among these works, Cellpose3D \cite{eschweiler2022robust} provides the best segmentation performance on the plant cells. Hence, this technique has been used in this paper. Cellpose3D learns the gradient map along the X,Y, and Z-axis. Using that gradient map and some post-processing steps, it predicts the instances of the plant cells. Fig. \ref{segmentation} shows the application of Cellpose3D to get the instances of plant cells.

\subsection{Registration}

 Cell growth and division cause physiological changes in the structure of the plant, which result in slight orientation and shift along X,Y and Z-axis. Also, during the live imaging technique, sometimes the plant has to be physically moved between different places. These incidents misalign the plant image stacks at different time points. 
 
 As plant cells are tightly packed, the positions of the cells do not change abruptly in a short time interval. Hence, if image stacks of plant are aligned (or registered) properly, we can focus on a small region of interest in the image stack, instead of searching the cell on the entire stack in order to track a particular cell. Thus, registration saves time and computational costs. Authors of \cite{mkrtchyan2013automated} proposed a landmark-based registration algorithm which obtained state-of-the-art results on registering two-dimensional slices. However, we note that the algorithm shows subpar performance for registration in three-dimensional space. This motivates us to propose a novel 3D registration algorithm, an overview of which has been shown in Fig. \ref{registration}. 
 
\begin{figure}
\centering
    \subfloat[]{{\includegraphics[width= 0.49 \columnwidth]{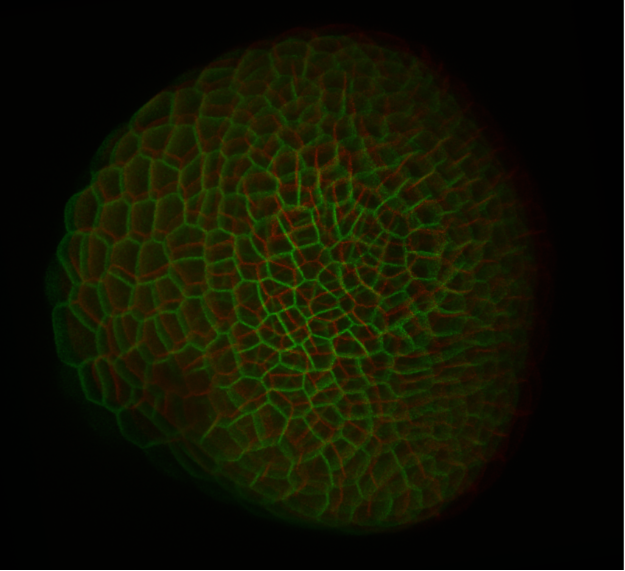}} \label{r1}}
    \hfill
    \subfloat[]{{\includegraphics[width= 0.49 \columnwidth]{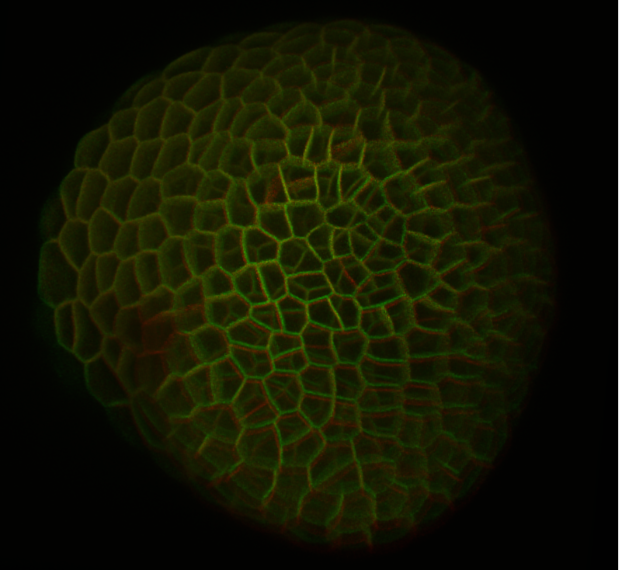}}\label{r2}}

    \caption{Overlay of cell borders of plant image stacks are shown. The red and green channels indicate cell borders of two different time points of SAM. (a) When plant image stacks are unregistered, cell borders of two-time points are visible separately. (b) After registration, the cell borders of two-time points get overlapped in most of the cases.}
    \label{reg_result}
\end{figure}

 Our proposed algorithm is a three step algorithm. The first step involves obtaining three-dimensional instance segmentation maps using pre-trained Cellpose3D models \cite{eschweiler2022robust}. The second step involves obtaining a number of pairwise cell correspondences (also known as seed pairs) using the algorithm proposed in \cite{liu2019deepseed}. Finally, Random Sample Consensus (RANSAC)  algorithm \cite{rusu2009fast} is used to determine the rotation and translation matrix from the point clouds of matching seed pair cells. The use of RANSAC is motivated by the fact that there can be some false positive matches in determining seed pairs in the second step and RANSAC is capable of estimating the parameters of a mathematical model from a set of matching data that contains false positive matches. The estimated rotation and translation matrix is used to perform registration between two plant image stacks.  Fig. \ref{reg_result} shows the overlays of two-time points from the top before  and after registration. However, in this registration technique, the plant is considered a rigid object which is not practically true. To address this issue we considered a region of interest for tracking cells described in the next sub-section.

\subsection{Cell Pair Matching}

It is evident that plant cells are non-rigid due to cell growth and cell division. Hence, the segmentation map of a cell gets deformed as it grows with time.  In addition, over-segmentation and under-segmentation may occur due to the presence of noise in the images. As a result, when we overlay the plant image stacks of two consecutive time points, the corresponding cells do not always overlap. To tackle this issue, we consider a region of interest to find a match for every cell. Let us consider a cell $s_{t_{1}}$ of time point $t_{1}$ which has centroid at coordinate $c_{t_{1}} \in \mathbb{R}^{3}$. There are total $q$ number of cells ${s^{(1)}_{t_{2}},s^{(2)}_{t_{2}}, \cdot \cdot \cdot,  s^{(q)}_{t_{2}} }$ in the next time point $t_{2}$ with their centroid coordinate at ${c^{(1)}_{t_{2}},c^{(2)}_{t_{2}}, \cdot \cdot \cdot,  c^{(q)}_{t_{2}} }$, respectively.

\begin{figure}
\centering
    \subfloat[]{{\includegraphics[width= \columnwidth]{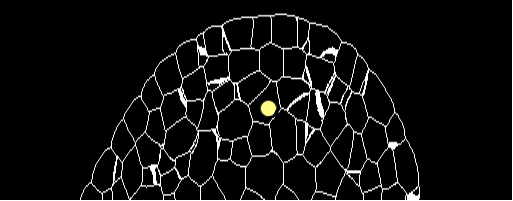}} \label{s1}}  
    
    \subfloat[]{{\includegraphics[width=\columnwidth]{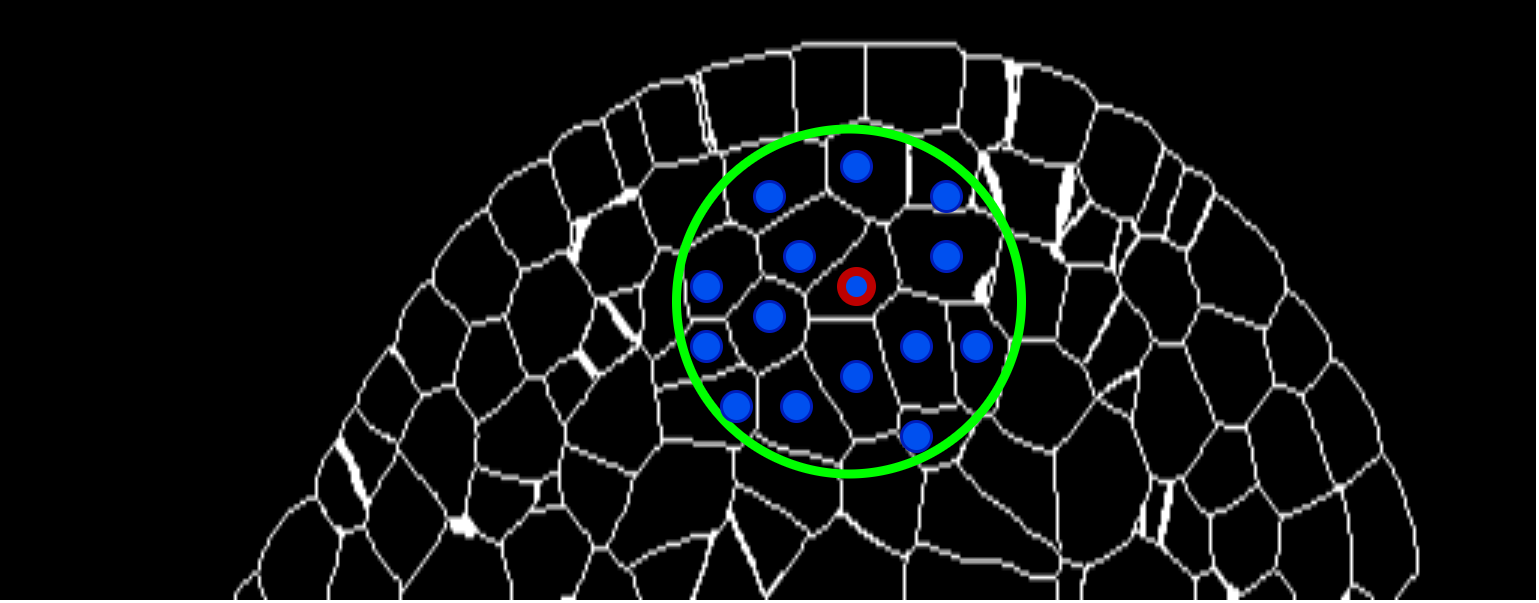}} \label{s2}}
    
    \caption{
Longitudinal section of plant for two consecutive time points where (a) and (b) indicate the former and later time point, respectively. In (a) cell with the yellow circle is the cell that needs to be tracked. In (b), cells marked by the blue circle are the candidates of matching and cell with a red border is the true matching. The green circle indicates our region of interest.}
\label{stack_cand}
\end{figure}

In order to find the corresponding cell of $s_{t_{1}}$ at time point $t_{2}$ , we consider $n$ number of candidate cells of set $\mathcal{R} = \{s^{(1)}_{t_{2}},s^{(2)}_{t_{2}}, \cdot \cdot \cdot,  s^{(n)}_{t_{2}}\}$ which have lowest distance from  coordinate $c_{t_{1}}$ among all other cells at time point $t_{2}$. Mathematically 

$$
\left\|c^{(1)}_{t_{2}}-c_{t_{1}}\right\| \leqslant\left\|c^{(2)}_{t_{2}}-c_{t_{1}}\right\| \cdots \leqslant\left\|c^{(n)}_{t_{2}}-c_{t_{1}}\right\|  
\cdots \leqslant\left\|c^{(q)}_{t_{2}}-c_{t_{1}}\right\|
$$
The region of interest comprises the cells in $\mathcal{R}$ as shown in Fig. \ref{stack_cand}. We use a 3D geometric feature extractor \cite{zhang2019deep}  and  propose a 3D graph matching algorithm to predict the corresponding cell of $s_{t_{1}}$ out of those candidates in $\mathcal{R}$.

\begin{figure}[h]
  \centering
  \includegraphics[width = \columnwidth]{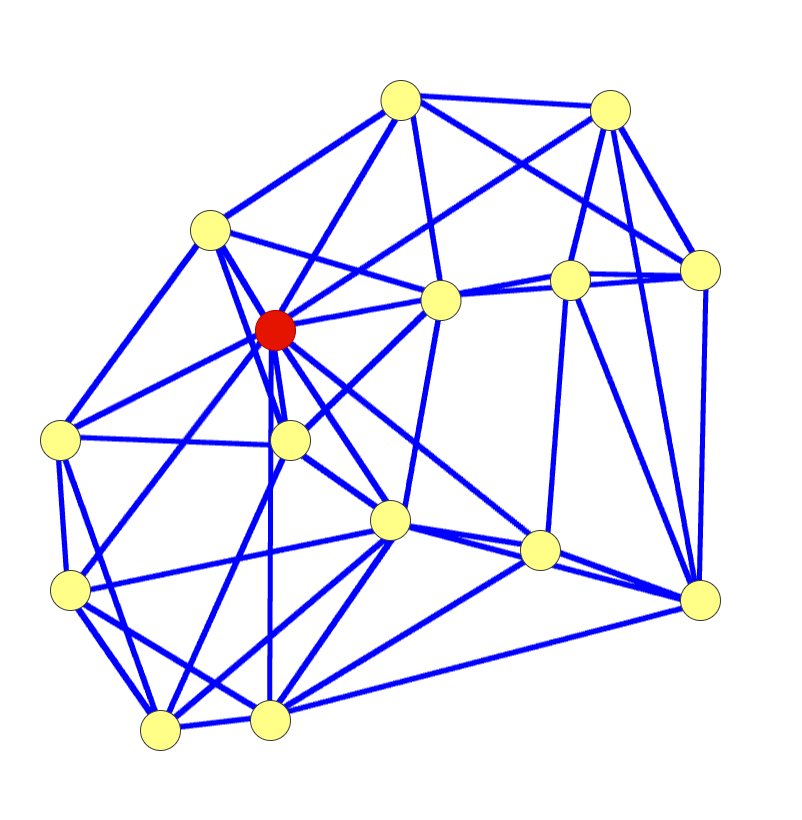}
  \caption{A three-dimensional graph generated by connecting the $k$-nearest neighbors. The central cell is shown by the red circle, and its neighboring cells are shown by yellow circles. Every cell is connected to its k-nearest neighbors (in the figure $k=5$ ).}
   \label{graph_one}
\end{figure}

% \begin{figure}
% \centering
%     \subfloat[] {{\includegraphics[width= 0.6\columnwidth]{images/pos.png}} \label{pos}}

%     \hfill
%     \subfloat[]{{\includegraphics[width=0.6\columnwidth]{images/semi_pos.png}}\label{semi_pos}} 
    
%     \subfloat[]{{\includegraphics[width=0.6\columnwidth]{images/neg.png}}\label{neg}}
    
%     \caption{This figure describes how the ground truth of the point-to-point supervision is set. The centroids of two cells which are considered for graph matching are overlayed on the red point. The blue and yellow points are the neighbors of those two cells, respectively. (a) the cells are matching cells and two local graphs are very similar to each other. Point-to-point supervision ground truth is 1 for all point pairs (shown in the figure). (b) the cells are matching cells. However, some point-pairs are far apart. The point pairs which have distances within a threshold value. have the ground truth $1$, otherwise, the ground truth is $0$. (c) the two cells are non-matching. All point-to-point ground truths are 0.       }\label{graph_gt}
% \end{figure}

\begin{figure*}
\centering
     % {{\includegraphics[width= \textwidth]{images/overlay.png}}} \label{overlay} \\

    \subfloat[] {{\includegraphics[width= 0.30\textwidth]{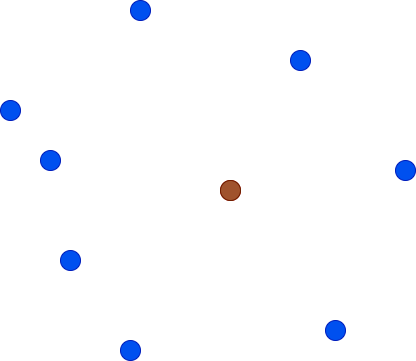}} \label{t0_g}}
    \hfill
    {{\includegraphics[width= 0.30\textwidth]{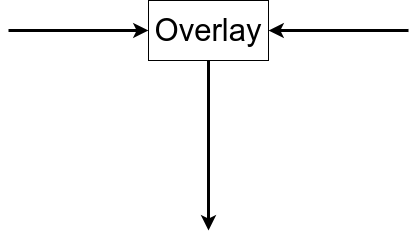}}} \label{ol}
    \hfill
    \subfloat[] {{\includegraphics[width= 0.30\textwidth]{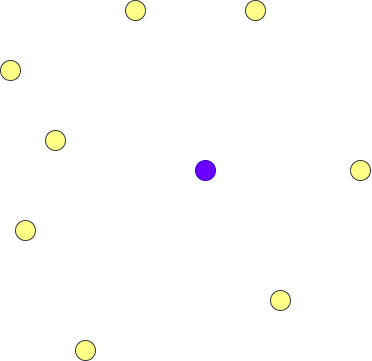}} \label{t1_g}}\\

    \subfloat[] {{\includegraphics[width= 0.30\textwidth]{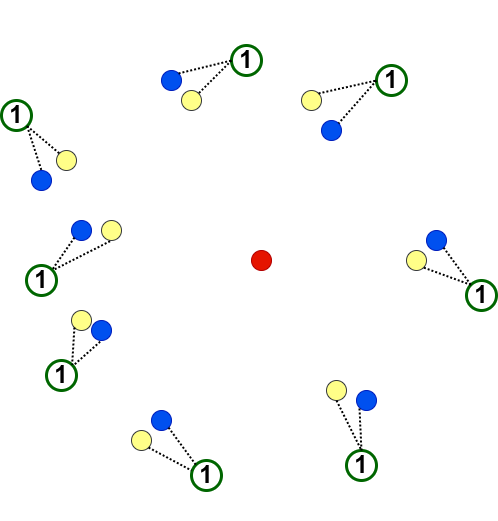}} \label{pos}}    
    \hfill
    \subfloat[]{{\includegraphics[width=0.3\textwidth]{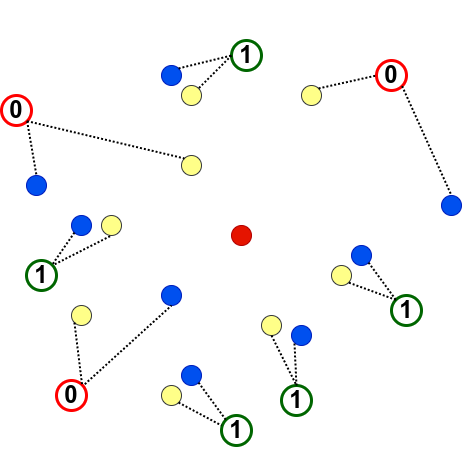}}\label{semi_pos}} 
    \hfill
    \subfloat[]{{\includegraphics[width=0.3\textwidth]{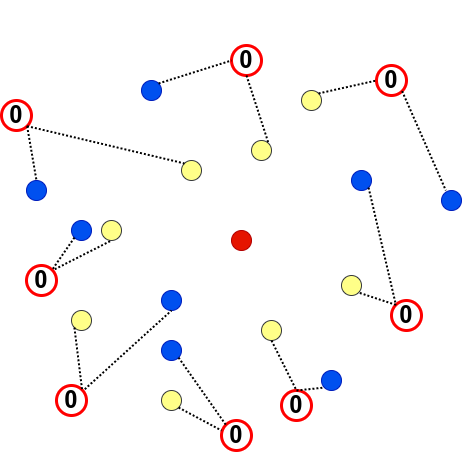}}\label{neg}}
    
    \caption{This figure describes how the ground truth of the point-to-point supervision is set. Every circle represents the centroid of a cell.  (a) the brown and blue circles represent the center of the graph and its neighbors, respectively at one time point. (b) the purple and yellow circles represent the the center of the graph and its neighbors, respectively at another time point. The centers of two graphs (brown and purple circles) are overlaid at the red color circle as shown in (c), (d) and (e). There can be three cases. Case - 1 :  (c) if the cells are matching cells, and two local graphs are less deformed, they are very similar to each other. Point-to-point supervision ground truth is set 1 for all point pairs (shown in the figure). Case - 2 : (d) if the cells are matching cells, but local graphs are slightly different due to deformation. In this case, some point-pairs are far apart. The point pairs which have distances within a threshold value is set the ground truth $1$, otherwise, the ground truth is set $0$. Case -3 : (e) if the two cells are non-matching, all the pairwise point-to-point ground truths are set to 0. }
    \label{graph_gt}
\end{figure*}

\subsubsection{3D Graph Matching}

The local graph for a particular cell comprises total $n_{p}$ nodes where the centroids of that cell and its $n_{p} - 1$ number of nearest neighbors in the three-dimensional space are considered. All these nodes are connected to $k$-nearest neighbors among themselves which forms the local 3D graph as shown in Fig. \ref{graph_one}. The formation of this type of graph helps us in two ways. Firstly, our graph is capable of extracting enriched features in three-dimensional space. The local graphs presented in \cite{liu2018multi,liu2019deepseed} consider two-dimensional star graph which mainly uses the positional information of neighboring cells with respect to the center cell only. On the other hand, our 3D local graph uses the positional information with respect to not only the central cell but also the neighboring cells among themselves which extracts rich three-dimensional spatial information. Secondly, our 3D local graph reduces computational cost compared to the graph proposed in \cite{chakraborty2015context} where local graph is not formed. Rather all the cells of a slice are considered as nodes of the graph, and all nodes corresponding to the cells having common boundaries are connected. However, the formation of this kind of large graph, makes the optimization computationally expensive and time-consuming. On the contrary, the 3D local graph is more inference friendly with less computational expense.

% For a  particular cell, the local 3D graph gets slightly deformed as time progresses. In order to track a particular cell we compare its local 3D graph with the local 3D graphs of the  potential matching candidates in the next time point. The candidate having the most similarity is considered as the matching cell. 

\begin{figure*}
  \centering
  \includegraphics[width = \textwidth]{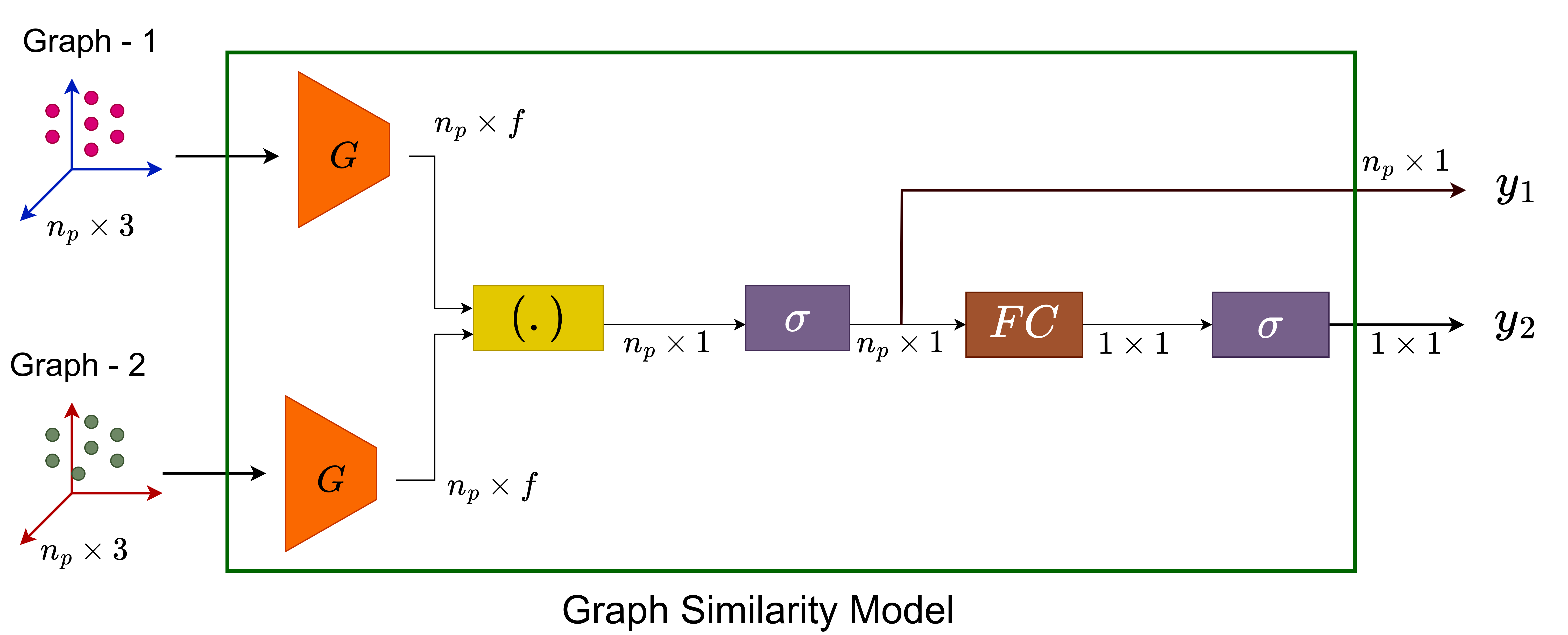}
  \caption{Graph Similarity Model (GSM) shown by the green box. There are two inputs of this model : three-dimensional coordinates of two graphs where each graph has $n_{p}$ number of points. The model provides two outputs. One is point-to-point similarity prediction ( $y_{1}$) for each point, and another is overall graph similarity prediction ( $y_{2}$). In the block diagram $G, (.), \sigma , \textit{FC}$ indicate geometric feature extractor, dot product, sigmoid activation function and fully connected layer, respectively.  In the figure input and output shape of every block is shown.}
  \label{graph_sim_model}
\end{figure*}

We propose a learning-based 3D graph matching approach where the input is an anchor graph (the 3D local graph of the cell we want to track) and $n$ number of  local 3D graphs of potential matching candidate cells in the next time point. The objective is to identify the candidate graph that exhibits the greatest similarity to the anchor graph. Fig. \ref{graph_sim_model} shows the graph similarity model where the inputs are the three-dimensional coordinates of two graphs. To extract the features of each node in a k-nearest neighbor connected graph, a geometric feature extractor \cite{zhang2019deep} is employed. The outputs of the model are the point-to-point similarity score (shown as $y_{1}$ )  overall graph similarity score(shown as $y_{2}$). As two graphs have the same number of nodes, we can make one-to-one association of nodes between two graphs based on their distance. Point-to-point similarity score indicates how similar the node pairs are based on their spatial orientation.  The overall graph similarity score indicates the overall similarity between two graphs.

\subsubsection{Training}

The 3D local graph matching is a fully supervised training. In this approach, joint supervision on both point-to-point similarity and overall graph similarity is needed. Point-to-point similarity supervision indicates pairwise node correspondence between two graphs. Fig. \ref{graph_gt} explains the point-to-point similarity supervision in detail. Overall graph similarity supervision indicates, out of $n$ candidate graphs which one is  the most similar to the anchor graph.  Mathematically, there are $n$ candidate graphs and $(y^{(1)}_{1},y^{(1)}_{2}), (y^{(2)}_{1},y^{(2)}_{2}), ......., (y^{(n)}_{1},y^{(n)}_{2})$ are the outputs of the networks, $p^{(1)},p^{(2)},......,p^{(n)}$ are the point-to-point similarity ground truth and $GT_{graph}$ is the overall graph similarity ground truth.

Binary cross-entropy loss ($BCE$) is used to train the point-to-point supervision. The point-to-point supervision loss ($L_{pp}$) can be expressed by

 \begin{equation}
\begin{split}
    L_{pp}  = \sum_{k_{1}=1}^{n} \sum_{k_{2}=1}^{n_{p}}BCE(y^{(k_{1})}_{1,k_{2}},p^{(k_{1})}_{k_{2}})
\end{split}
\end{equation}

We note that there are some cases where cell deformation or segmentation errors occur very much. In those cases, the region of interest shown in Fig. \ref{stack_cand} does not always include the true matching cell. Hence we considered an extra class $y^{'}$ for graph similarity supervision which indicates the anchor graph does not match with any candidate graph. The score for $y^{'}$ can be expressed by 

\begin{equation}
    y^{'}  = 1 - \max(y^{(1)}_{2},y^{(2)}_{2},...,y^{(n)}_{2})
\end{equation}

 Hence, the anchor graph can match with any of the members of set $Y_{cand} = \{y^{(1)}_{2},y^{(2)}_{2}, \cdot \cdot \cdot,  y^{(n)}_{2}, y^{'}\}$ which is guided by cross-entropy (CE) loss. The overall graph similarity supervision ($L_{g}$) loss can be expressed by 
 
 % $Y_{cand} =  \{y^{(1)}_{2},y^{(2)}_{2}, \cdot \cdot \cdot \cdot,y^{(n)}_{2}, y^{'}}\}$
 % . Either the anchor graph can match with any of the $n$ candidate graphs, or there can be no matching with candidate graph. 

%  \begin{equation}
% \begin{split}
%     L_{g}  = CCE(GT_{graph},y^{(1)}_{2},y^{(2)}_{2},...,y^{(n)}_{2}, y^{'})
% \end{split}
% \end{equation}

 \begin{equation}
\begin{split}
    L_{g}  = CE(GT_{graph},Y_{cand})
\end{split}
\end{equation}

Now the total loss can be written as

\begin{equation}
\begin{split}
    L_{total}  = \lambda L_{pp} + L_{g}
\end{split}
\end{equation}

where $\lambda$ is a non-negative hyper-parameter that controls the weight of point-to-point similarity supervision in the joint loss function.

\begin{figure}
  \centering
  \includegraphics[width = 0.5 \textwidth]{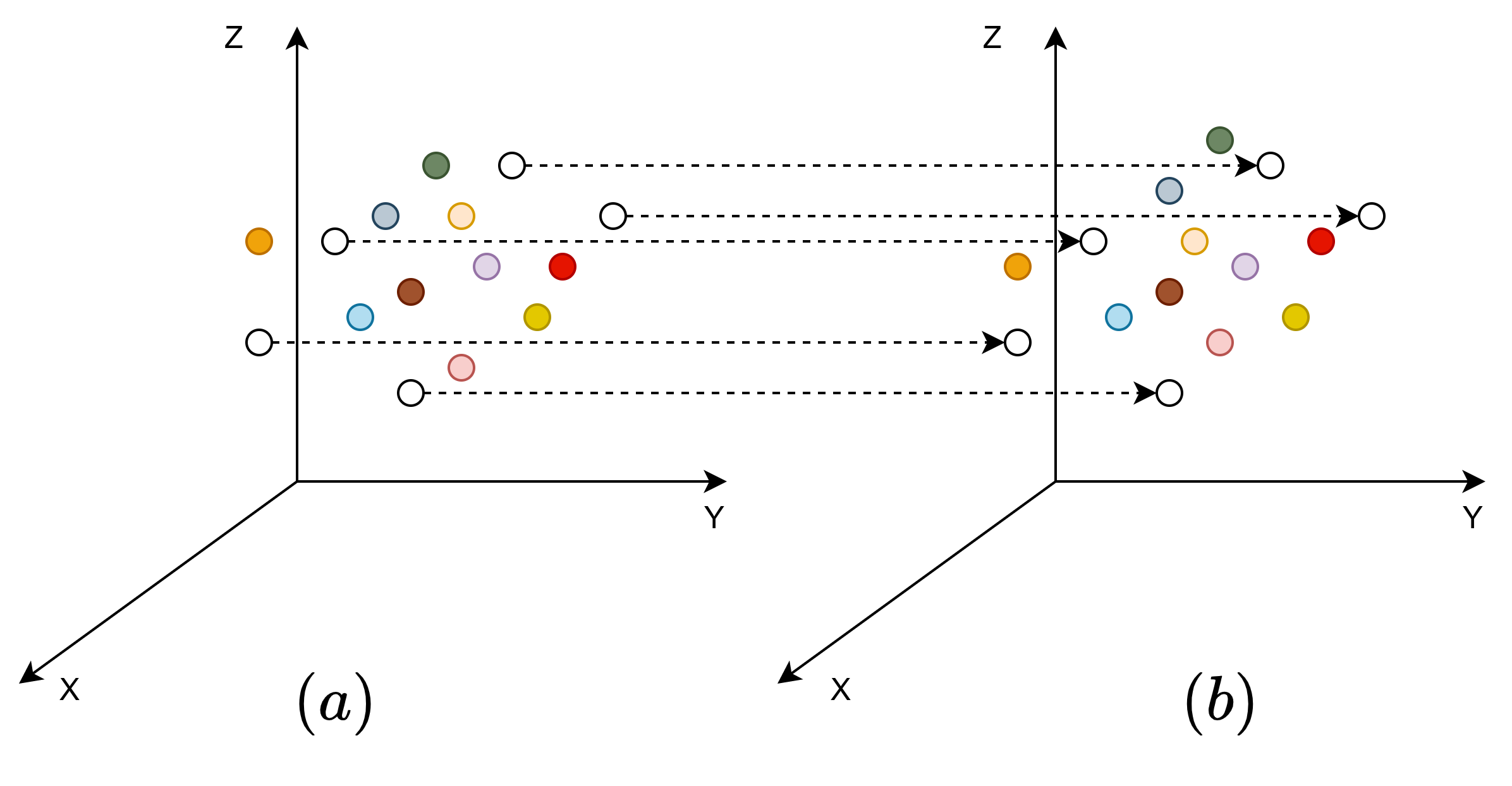}
  
  \caption{(a) and (b) indicate the cells of former and later time point, respectively. The circles indicate centroids of the cells. Using the learning-based tracking method most of the cells are tracked. The same coloured cells indicate the matching pairs. However, the white colour circles indicate the cells which could not be tracked using our method. Using the relative positions of the tracked cells (coloured), we can associate the untracked cells (shown by the dotted arrow).}
  \label{track_rest}
 
\end{figure}

\begin{figure*}
  \centering
  \includegraphics[width = \textwidth]{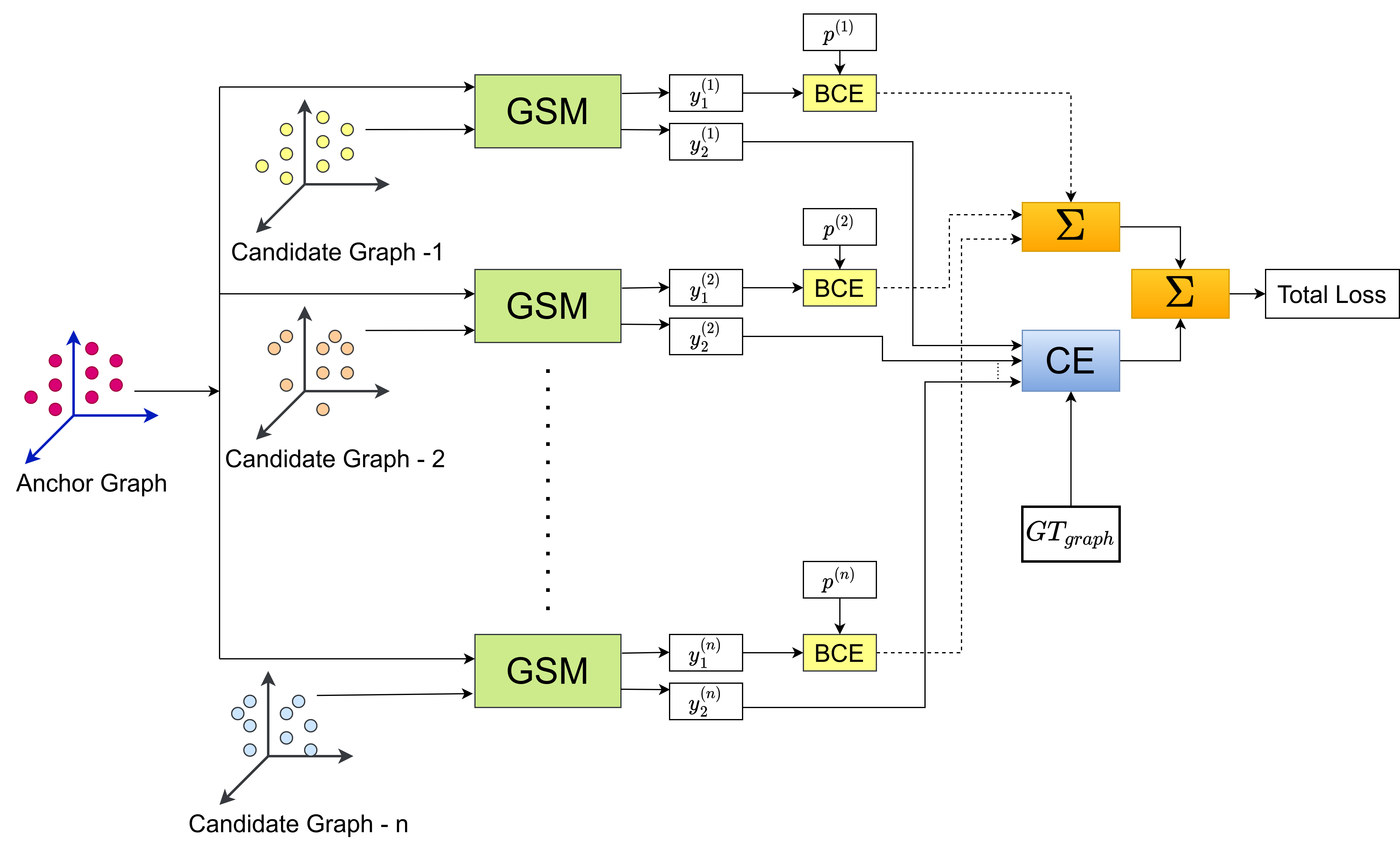}
  \caption{The figure shows that the anchor graph is compared against $n$ candidate graphs. For each candidate, the anchor graph and the candidate graphs are inputs of GSM (Graph Similarity Model as shown in Fig. \ref{graph_sim_model}). For each model there are two outputs - point-to-point similarity prediction ($y^{(k)}_{1}$ where k = 1,2, \ldots ,n) and overall graph similarity ($y^{(k)}_{2}$ where k = 1,2, \ldots ,n) prediction which are guided by the BCE loss and CE loss, respectively.  }
  \label{graph_sim_train}
\end{figure*}

Fig. \ref{graph_sim_train} demonstrates the overall training procedure. Although this approach can find most of the matching cell pairs, some cells remain unmatched. Inspired by \cite{liu2019deepseed}, we can track the unmatched cells using their relative positions with respect to matching pair cells (demonstrated in Fig. \ref{track_rest}).

\begin{figure*}
    
    \centering
    \includegraphics[width = 0.9\textwidth]{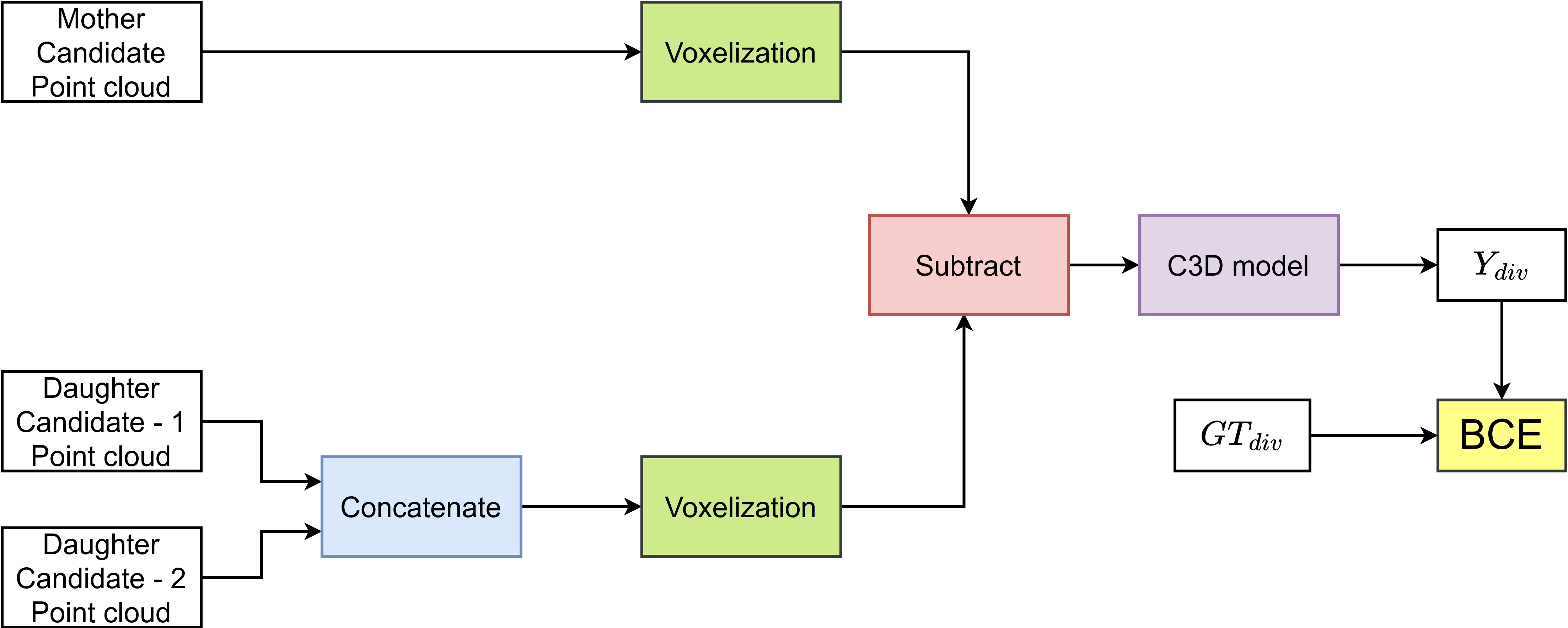}
    \caption{Three-dimensional shape and volume similarity prediction between mother and combined daughter cells. Both the point cloud of mother cell and concatenated daughter cell pair is voxelized and subtracted. The subtracted result is trained by 3D convolutional model (C3D model). The training is guided by Binary Cross Entropy (BCE) loss.}
    \label{cell_div_dgm}
 
\end{figure*}

\subsection{Cell Division Detection}

Cell division \cite{mazia1961mitosis}  is the process by which a mother cell gets divided into two daughter cells. The state-of-the-art cell division detection techniques \cite{basu2011mapping}, \cite{liu2017multi} were based on two assumptions: \begin{itemize}
    \item The combined area of daughter cells is almost equal to the area of mother cell.
    \item The combined cell borders of daughter cells have an almost similar shape to the mother cell's border.
\end{itemize}  

However, the outcome from these assumptions is highly dependent on the accuracy of segmentation and does not consider cell growth over time. In this paper, we have introduced a novel learning-based method for detecting cell division. This method uses two sources of information to detect cell division: the 3D shape similarity between mother and daughter cells and the 3D local graph similarity between mother and daughter cells. For every potential mother cell, a region of interest (same as \ref{stack_cand}) is selected in the next time point, and potential daughter cell pairs are selected by taking pairs of adjacent cells in that region of interest. 

\subsubsection{3D Shape and Volume Similarity}
In three-dimensional space, the shape and volume of combined daughter cells are almost equal to the the shape and volume of mother cell. In order to use this physical property, we present a workflow shown in Fig. \ref{cell_div_dgm} which is totally data-driven.
Voxelization \cite{chen2020every} is done on the point clouds of mother and daughter cells. Then subtraction is done between the voxelized mother cell and concatenated voxelized daughter cells. The subtracted result represents distinct patterns based on whether the input point clouds are from mother-daughter cell pairings or not. A 3D convolutional neural network \cite{tran2015learning} is trained to do binary classification between these two separate patterns, which is guided by a BCE loss. 
$Y_{div}$ is the output of the model which is a probability score on how similar the mother and daughter pairs are in terms of three-dimensional shape. $GT_{div}$ is the ground truth for cell division. If the inputs are true mother-daughters $GT_{div}$ is 1, otherwise 0. 

\begin{equation}
     L_{\text{cell sim}}  = BCE(Y_{div},GT_{div})
     \label{eq_loss_cell_sim}
\end{equation}

\subsubsection{Pairwise Local Graph Similarity}
Spatially the mother and daughter cells occupy almost at the same location of the plant. Hence the 3D local graph with respect of mother cell and each of the daughter cells are almost similar. In order to quantify, how similar two local 3D graph are, we use the network presented in Fig. \ref{graph_sim_model}. The loss function in order to determine pairwise local graph similarity can be expressed by    

\begin{equation}
     L_{\text{graph pair}}  = \sum_{k=1}^{n_{p}}BCE(y_{1,k},p_{k}) + BCE(y_{2},GT_{\text{graph pair}})
     \label{eq_loss_graph_pair}
\end{equation}

$GT_{\text{graph pair}}$ is the ground truth for training graph pair similarity. It is set to 1 if two graphs are similar, otherwise 0. The overall graph similarity score $y_{2}$ predicts a score based on the similarity of 2 local 3D graph inputs. 

Let $m$ is the mother and $({d_{1},d_{2}})$ are daughter pairs. For each of the inputs of $(m,d_{1})$ and $(m,d_{2})$ the network of Fig. \ref{graph_sim_model} predicts overall graph similarity score $y_{2,d_{1}}$ and $y_{2,d_{2}}$ ,respectively ,the cell division score can be expressed by

 % We use the network of  in order to  get a probability score on how similar the local graphs of the mother and each of the daughter cells are. In this case both $y_{1}$ and $y_{2}$ are guided by BCE loss.

% Here $y_{2}$ is pairwise graph similarity score and $GT_{graph}$ is the ground truth on similarity.  $GT_{graph}$ is 1 if two graphs are similar, otherwise 0.

\begin{equation}
     Score_{\text{div}}  = w_{d}Y_{div} + (1-w_{d})\frac {(y_{2,d_{1}} + y_{2,d_{2}})}{2}
     \label{eq_cell_div_score}
\end{equation}

 where $w_{d}$ is a weighting factor for cell division which is set to $0.5$. If the value of $Score_{\text{div}}$ is higher than a threshold, then it is reasonable to assume that cell division occurred.

\section{Experiments and Results}
\label{sec:Experiment}

\subsection{Dataset}

For the experiments, we have used publicly available confocal imaging based plant cell dataset \cite{willis2016cell} consisting of six plants. For each plant, on average 20 time points data were provided with a gap of four hours between two consecutive time points. Each time point image stack has around 200 slices and each slice has the size $512 \times 512$ pixel.  Two types of ground truth are provided with this dataset. One is for segmentation and another is for tracking. In segmentation ground truth, instance segmentation of each cell was given. In the tracking ground truth, pairwise cell correspondence is provided between two-time points. 

\subsection{Experimental Settings}
For cell tracking, out of the available six plant data, four plants have been used for training, one for validation and one for testing. We also engaged cross-validation to avoid bias. From the recent 3D segmentation techniques such as spherical harmonics \cite{eschweiler2021spherical} , Cellpose3D \cite{eschweiler2022robust}, we have used Cellpose3D. The spherical harmonics approach does not preserve the polygonal shape of the cell which makes this approach inappropriate to detect cell division. On the other hand, Cellpose3D preserves the polygonal shape of the plant cell which makes this approach suitable for our work. The pre-trained model of Cellpose3D is publicly available and according to \cite{eschweiler2022robust}, experiments on segmentation was done on  the same dataset with the same data-split we have used.

\subsection{Results}

% \begin{figure*}
%   \centering
%   \includegraphics[width = 0.7 \textwidth]{images/tra.png}
  
%   \caption{(a) and (b) indicates the cells of former and later time point, respectively. The same coloured cells of these figures indicate the corresponding cell pairs.}
%   \label{cell_tra_result}
 
% \end{figure*}

% \begin{figure}
%   \centering
%   \includegraphics[width = \columnwidth]{images/tra.png}
  
%   \caption{(a) and (b) indicates the cells of former and later time point, respectively. The same coloured cells of these figures indicate the corresponding cell pairs.}
%   \label{cell_tra_result}
 
% \end{figure}

\begin{figure}
\centering
    \subfloat[]{{\includegraphics[width= 0.49 \columnwidth]{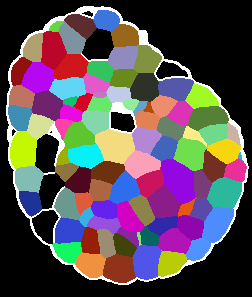}} \label{tr_0}}
    \hfill
    \subfloat[]{{\includegraphics[width= 0.49 \columnwidth]{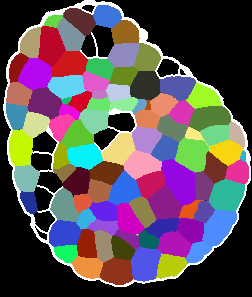}}\label{tr_1}}

  \caption{(a) and (b) indicates the tracked cells of former and later time point, respectively. The same coloured cells of these figures indicate the matching cell pairs.}
  \label{cell_tra_result}
\end{figure}

To assess the tracking performance of our proposed approach, we conduct a comparison with the method described in \cite{chakraborty2015context}. \footnote{We wanted to compare our method against two recent papers \cite{liu2019deepseed} and \cite{deep_patch}. The authors of those papers have not made their codes available. We were not able to reproduce their results by our own implementation. In addition, the results reported in their papers are for their private dataset.} This method has used watershed segmentation \cite{beucher1993morphological} which basically segments the cell border of the plant cell. In our work, we have used Cellpose3D for segmentation. While comparing with \cite{chakraborty2015context}, we have used the cell borders from the segmentation provided by Cellpose3D  instead of using noisy watershed segmentation in order to make a fair comparison. In addition, the method described in \cite{chakraborty2015context} uses landmark-based 2D registration \cite{mkrtchyan2013automated}, while we use 3D registration. Hence, we have provided the comparative 
results for both 2D and 3D registration.

\subsubsection{Performance of Cell Pair Matching}

We present a comparison of our novel method with \cite{chakraborty2015context} in TABLE \ref{table_overall} for pairwise cell matching. The comparison encompasses both two-dimensional registration and the newly introduced three-dimensional registration. The evaluation indices are precision, recall and F1 score which can be expressed by the following equation. 

\begin{equation}
\text { Precision }=\frac{T P}{T P+F P}
\end{equation}

\begin{equation}
\text { Recall }=\frac{T P}{T P+F N}
\end{equation}

\begin{equation}
F 1 \text { score }=\frac{2 \times \text { Precision } \times \text { Recall }}{\text { Precision }+\text { Recall }}
\end{equation}

where TP, FP and FN indicate true positive, false positive and false negative respectively.

From TABLE  \ref{table_overall}, we see that our proposed method shows better performance than other compared methods. We also note that 3D registration has improved the performance of the method described in \cite{chakraborty2015context} compared to 2D registration. Overall, our method has secured $6.83 \%, 5.96 \% \text{and}, 6.40 \% $ improvement in terms of precision, recall and F1 score, respectively when 3D registration is considered. Fig. \ref{cell_tra_result} shows some visual results of cell pair matching using our method.

\begin{table}

\centering
\caption{Performance comparison among different methods of pairwise cell matching.}
\label{table_overall}

{\small
\begin{tabularx}{\columnwidth} 
{  c c c c c }

Method      & Regis. & Precision       &Recall   &F1 sc.  \\
\hline

Chakraborty {\it et al.} \cite{chakraborty2015context} & 2D          &0.7645             &0.8118                 &0.7874 \\

Chakraborty {\it et al.} \cite{chakraborty2015context} & 3D          &0.9209            &0.8992                &0.9099 \\

Ours    & 3D      &\textbf{0.9892}              &\textbf{0.9588}                  &\textbf{0.9739} \\
% Ours          &0.99596              &0.97379                  &0.98475 \\

\end{tabularx}
}

\end{table}

% \begin{table}

% \centering
% \caption{Performance comparison among different methods of plant cell tracking.}
% \label{table_overall}

% {\small
% \begin{tabularx}{\columnwidth} 
% {  c|c| c c c  }
%  \hline
% Method      & Registration &Precision       &Recall   &F1 sc.  \\
% \hline

% Chakraborty {\it et al.} \cite{chakraborty2015context} & 2D          &0.7645             &0.8118                 &0.7874 \\

% Chakraborty {\it et al.} \cite{chakraborty2015context} & 3D          &0.9209            &0.8992                &0.9099 \\

% Method - 2 \cite{liu2019deepseed}   & 2D        &0.3360              &0.4827                 &0.3962 \\
% Method - 2 \cite{liu2019deepseed}   & 3D        & 0.4461              & 0.5910                 & 0.5085  \\

% Ours    & 3D      &\textbf{0.9892}              &\textbf{0.9588}                  &\textbf{0.9739} \\
% % Ours          &0.99596              &0.97379                  &0.98475 \\
% \hline

% \end{tabularx}
% }

% \end{table}

\subsubsection{Performance on cell division}

\begin{figure}
  \centering
  \includegraphics[width =  \columnwidth]{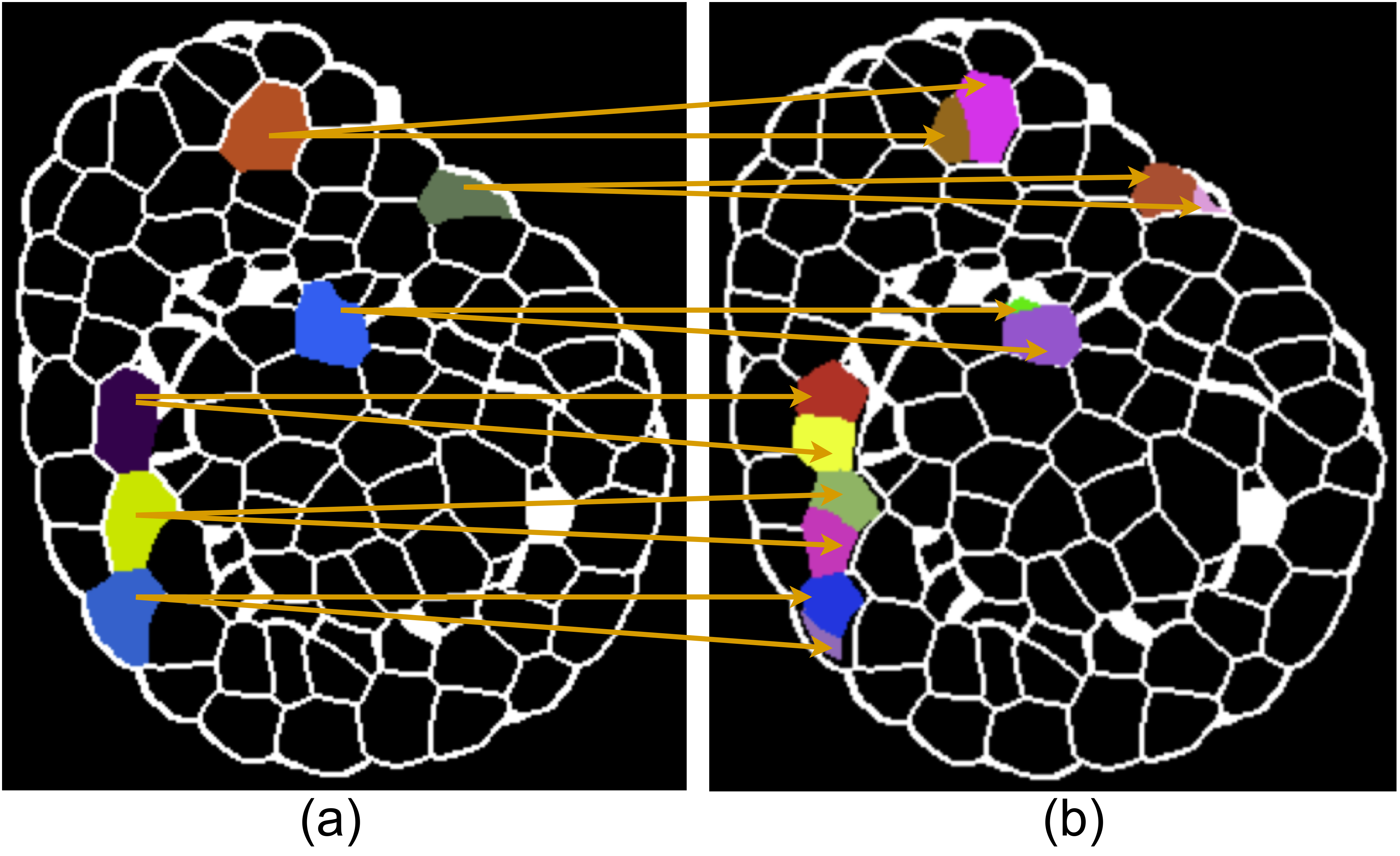}
  
  \caption{Coloured cells of (a) indicate the mother cells and coloured cells of (b) indicate the  daughter cells. The arrow sign shows mother-daughter correspondence. }

  \label{cell_div_result}
 
\end{figure}

TABLE \ref{table_division} shows the comparison of performances of our novel method with   \cite{chakraborty2015context} for cell division. It is evident that our method outperforms the \cite{chakraborty2015context} ( 2D registration ) in all metrics. When 3D registration is considered for \cite{chakraborty2015context}, our method shows competitive result on precision and shows $15.38 \%$ and $14.78 \%$ improvement in recall and F1-score, respectively. Fig. \ref{cell_div_result} some visual results of cell division using our method.

\begin{table}

\centering
\caption{Performance comparison among different methods for cell division.}
\label{table_division}

{\small
\begin{tabularx}{\columnwidth} 
{  c c c c c }

Method      & Regis. & Precision       &Recall   &F1 sc.  \\
\hline

Chakraborty {\it et al.} \cite{chakraborty2015context} & 2D          &0.9302              &0.2857                  &0.4372 \\

Chakraborty {\it et al.} \cite{chakraborty2015context} & 3D          & \textbf{1.0000}             &0.3521                 &0.5208 \\

% Method - 2 \cite{liu2019deepseed}   & 2D        &0.9437              &0.1588                  &0.2718 \\
% Method - 2 \cite{liu2019deepseed}   & 3D        & 0.9638             & 0.2419                 & 0.3867  \\

Ours    & 3D      & 0.9892              & \textbf{0.5059}                 & \textbf{0.6686} \\

\end{tabularx}
}

\end{table}

\subsubsection{Cell lineage generation}
Traditionally multiple object tracking (MOT) algorithms \cite{kim2018multi,voigtlaender2019mots} rely on tracking-by-detection. In these approaches, the input is single frame of image. However, our proposed method takes a pair of image stacks of consecutive time points as inputs. Then pairwise cell matching and cell division are detected. The pairwise matching and division can be considered as a small tracklet. Combining the small tacklets can create cell linages in the long term tracking process. Motivated by \cite{peng2020chained}, Xie et al. \cite{deep_patch} proposed an efficient node chaining method to generate plant cell trajectories on multiple time points. We have used the save method to generate long term cell linage. Table \ref{table_trajectory} shows the comparisons of the accuracy among three methods in terms of determining long-time cell lineages. We note that dataset \cite{willis2016cell}, does not provide long-time linages of all the cells. In the evaluation of long time tracks, we have only considered those cells which have ground truth tracks of at least 16 hours. The accuracy was measured based on the average percentage of experimental linage matching with respect to the ground truth lineages. Our proposed method outperforms the other two methods by $23.0262 \%$ , $12.964 \%$, respectively .

% \begin{table}
% \centering
% \caption{Performance comparison on determining long-time trajectory accuracy}
% \label{table_trajectory}

% {\small
% \begin{tabularx}{\columnwidth} 
% { | c | c |}
%  \hline
% Method      & Accuracy  \\
% \hline

% Chakraborty {\it et al.} \cite{chakraborty2015context}   & 50.9012 \%\\
% \hline
% Method - 2 \cite{liu2019deepseed}           & 25.7087 \%\\
% \hline
% Ours          &59.6373 \% \\
% % Ours          &0.99208              &0.96558                  &0.97865 \\

% \hline

% \end{tabularx}
% }

% \end{table}

\begin{table}
\centering
\caption{Performance comparison on determining long-time trajectory accuracy}
\label{table_trajectory}

{\small
\begin{tabularx}{\columnwidth} 
{ c  c  c }

Method      & Regis. & Accuracy (\%) \\
\hline

Chakraborty {\it et al.} \cite{chakraborty2015context}   & 2D & 47.9631 \\
% \hline
Chakraborty {\it et al.} \cite{chakraborty2015context}   & 3D & 58.2929 \\
% \hline
Ours          &3D &\textbf{70.9893}  \\

\end{tabularx}
}

\end{table}

\subsubsection{Performance on running time}

Our novel approach has done huge improvement in terms of running time. The inference time of our method is around $19$ hours while method \cite{chakraborty2015context} takes around $236$ hours. Hence our method is around $12$ times faster.

\subsection{Ablation Study}
The proposed cell pair matching method has two steps. In the first step, we find pairwise cell matching using a deep neural network. However, some cells are left unmatched in this step. In the second step, we track those untracked cells using their relative position with respect to the tracked cells. Clearly, in the second step, no learning-based approach is involved. In our ablation study, we show the comparison between one-step and two-step tracking while changing different factors of tracking such as the number of connections to neighbors in graph formation, the size of the region of interest and the impact of modified loss function during training.

\subsubsection{Effect of changing the number of connections with neighbors in graph formation}

Given the centroids of $n_{p}$ number of cells, we form the graph by joining $k$ nearest points among themselves.  In TABLE \ref{table_k_change} we show how the results of cell matching change with different values of $k$. We note that with the increase of $k$, the precision value increases. Because an increase of $k$ means more information is used regarding the spatial orientation of the nodes. Hence cell matching accuracy increases with the  increase of $k$. In addition, while using one-step tracking, some cells are left untracked. In two-step tracking, some of those untracked cells can find their matching pairs. As a result, we observe an increase of recall value in two-step tracking.

\begin{table}

\centering
\caption{Effect of changing the number of collections with neighbors ($k$)}
\label{table_k_change}

{\small
\begin{tabularx}{\columnwidth} 
% { |c| p{10 mm} p{10 mm} p{10 mm} | p{10 mm} p{10 mm} p{10 mm}|  }
{ c| c c c| c c c  }

 &\multicolumn{3}{c|}{\text { One step tracking}} & \multicolumn{3}{c}{\text { Two step tracking }} \\
 \hline

$k$     &Prec.       &Recall   &F1 sc. &Prec.      &Recall   &F1 sc. \\

\hline
4   &0.9801      &0.9074      &0.9423   &0.9781      &0.9567      &0.9673\\

5   &0.9880      &0.9130      &0.9491   &0.9804      &0.9575      &0.9688\\

6   &0.9909      &0.9219      &0.9551   &0.9892      &0.9588      &0.9739\\

\end{tabularx}
}

\end{table}

\subsubsection{Effect of changing the size of region of interest}
The size of region of interest is related to  $n$ which indicates how many candidate graphs are compared against the anchor graph. TABLE \ref{table_n_change} shows when $n$ increases, the precision value decrease while the recall value increases in one-step tracking. It is because when  $n$ is lower, the network learns to classify for lower number of classes which is easier. Hence the precision value gets higher with low value of $n$ is used. On the other hand with the lower value of $n$, the region of interest does not always cover the true matching cell. As a result, the number of unmatched cells increases which lowers the recall value.  

\begin{table}

\centering
\caption{Effect of changing the number of matching candidates ($n$)}
\label{table_n_change}

{\small
\begin{tabularx}{\columnwidth} 
% { |c| p{10 mm} p{10 mm} p{10 mm} | p{10 mm} p{10 mm} p{10 mm}|  }
{ c| c c c| c c c  }

 &\multicolumn{3}{c|}{\text { One-step tracking}} & \multicolumn{3}{c}{\text { Two-step tracking }} \\
 
\hline
$n$     &Prec.       &Recall   &F1 sc. &Prec.       &Recall   &F1 sc. \\
\hline

5   &0.9978      &0.8887     &0.9401   &0.9937      &0.9694      &0.9814\\

% 10   &0.9839      &0.8771      &0.9274   &0.9781      &0.9567      &0.9673\\
% \hline
10   &0.9858      &0.9058      &0.9441   &0.9819      &0.9503      &0.9658\\

20   &0.9909      &0.9219      &0.9551   &0.9892      &0.9588      &0.9739\\

\end{tabularx}
}

\end{table}

\begin{table}[!h]

\centering
\caption{Impact of individual loss terms}
\label{table_loss_change}

{\small
\begin{tabularx}{\columnwidth} 
% { |c| p{10 mm} p{10 mm} p{10 mm} | p{10 mm} p{10 mm} p{10 mm}|  }
{ c| c c | c c  }

 &\multicolumn{2}{c|}{\text { One-step tracking}} & \multicolumn{2}{c}{\text { Two-step tracking }} \\
 
\hline
Loss      &Prec.       &Recall   &Prec.       &Recall \\
\hline

$L_{pp}$   &0.9408     &0.3605        &0.8305      &0.8819 \\

$L_{g}$   &0.9394      &0.6928        &0.9316      &0.9132 \\

$L_{pp}+L_{g}$   &0.9909      &0.9219         &0.9892      &0.9588 \\

\end{tabularx}
}

\end{table}

% \begin{figure}[!htb]
%   \centering
%   \includegraphics[width = \columnwidth]{images/longer_interval.eps}
  
%   \caption{}
%   \label{longer_interval}
 
% \end{figure}

% \subsubsection{Cell pair matching performance in longer time interval}
% In our dataset, the data was recorded with a constant interval 4 hours. Here we evaluate how the network performs in cell pair matching with longer time intervals. 

\subsubsection{Effect of individual loss terms}
The learning-based 3D graph matching was guided by two losses. They are point-to-point similarity supervision loss ($L_{pp}$) and overall graph similarity supervision loss ($L_{g}$). In TABLE \ref{table_loss_change} , we demonstrate the importance of both of the loss terms. If we train the network with only one loss term, all the metrics show lower values in both one and two-step tracking. Hence, both of the loss terms are necessary for effective cell tracking.

\section{Conclusion}

\label{sec:Conclusion}

In this paper, we have proposed a novel learning-based method to automatically track plant cells in any kind of three-dimensional unregistered condition. Unlike the state-of-the-art methods where tracking was done by two-dimensional graph matching, our proposed method constructs a three-dimensional graph that extracts the tight spatial features in a better way taking cell growth, deformity, and segmentation error under consideration. We also proposed a novel learning-based cell division detection method that  has performed better than the existing methods. Our method alleviates the limitations of state-of-the-art methods by improving accuracy and reducing running time which makes this approach more inference friendly.

\ifCLASSOPTIONcompsoc
  \section*{Acknowledgments}
\else
  \section*{Acknowledgment}
\fi

The authors would like to thank...

\ifCLASSOPTIONcaptionsoff
  \newpage
\fi



%\bibliographystyle{IEEEtran}

\bibliographystyle{IEEEtran}
\bibliography{IEEEabrv,Bibliography}


% H.~Kopka and P.~W. Daly, \emph{A Guide to {\LaTeX}}, 3rd~ed.\hskip 1em plus

% \end{thebibliography}


% \end{IEEEbiography}


%\newpage



%\vfill


\end{document}